\pdfoutput=1
\documentclass[10pt,twocolumn,letterpaper]{article}
\usepackage[pagenumbers]{cvpr}
\usepackage{graphicx}
\usepackage{amsmath,amssymb}
\usepackage{booktabs}
\usepackage{xcolor}
\usepackage{microtype}

\usepackage{colortbl}
\definecolor{rankwarm}{HTML}{B65338}
\newcommand{\rankfirst}[1]{\cellcolor{rankwarm!52}\textbf{#1}}
\newcommand{\ranksecond}[1]{\cellcolor{rankwarm!36}#1}
\newcommand{\rankthird}[1]{\cellcolor{rankwarm!24}#1}
\newcommand{\ranklegend}{\begingroup\setlength{\fboxsep}{1.5pt}%
  \colorbox{rankwarm!52}{\strut\textbf{top-1}}\,/%
  \colorbox{rankwarm!36}{\strut top-2}\,/%
  \colorbox{rankwarm!24}{\strut top-3}\endgroup}

\usepackage{cuted,longtable,float}
\definecolor{paperlink}{HTML}{95513E}
\usepackage[breaklinks,colorlinks,allcolors=paperlink]{hyperref}
\makeatletter
\g@addto@macro\UrlBreaks{\do\0\do\1\do\2\do\3\do\4\do\5\do\6\do\7\do\8\do\9}
\makeatother
\newcommand{\PreprintAuthors}{%
  Chen Xun\textsuperscript{1,*}\quad Xuefeng Chen\textsuperscript{2,*}\quad Linbo Zhai\textsuperscript{3,*}\quad Bin Wang\textsuperscript{2}\\
  Likui Zhang\textsuperscript{4}\quad Tianle Zhang\textsuperscript{5}\quad Jiaming Liang\textsuperscript{5}\quad Tianchen Deng\textsuperscript{6,\textdagger}\\[3pt]
  {\small \textsuperscript{1}Nanyang Technological University\quad \textsuperscript{2}Tsinghua University\quad \textsuperscript{3}South China University of Technology}\\
  {\small \textsuperscript{4}Sun Yat-sen University\quad \textsuperscript{5}JD.com\quad \textsuperscript{6}Shanghai Jiao Tong University}\\
  {\small \textsuperscript{*}Equal contribution.\qquad \textsuperscript{\textdagger}Corresponding author.}%
}
\newcommand{\PreprintPDFAuthors}{Chen Xun, Xuefeng Chen, Linbo Zhai, Bin Wang, Likui Zhang, Tianle Zhang, Jiaming Liang, Tianchen Deng}

\title{GT-VLM: Learning to Reason and Act with Geometry Tokens}
\author{\PreprintAuthors}
\hypersetup{pdftitle={GT-VLM: Learning to Reason and Act with Geometry Tokens},pdfauthor={\PreprintPDFAuthors}}
\makeatletter
\renewcommand{\@maketitle}{\begin{center}
  {\Large\bfseries\@title\par}\vspace{8pt}
  {\large\begin{tabular}{@{}c@{}}\@author\end{tabular}\par}
\end{center}}
\makeatother
\newcommand{\supppair}[2]{\par\smallskip\noindent
  \begin{minipage}[t]{0.48\linewidth}\vspace{0pt}#1\end{minipage}\hfill
  \begin{minipage}[t]{0.48\linewidth}\vspace{0pt}#2\end{minipage}\par\smallskip}

\begin{document}
\maketitle
\begin{strip}
\centering
\includegraphics[width=\textwidth]{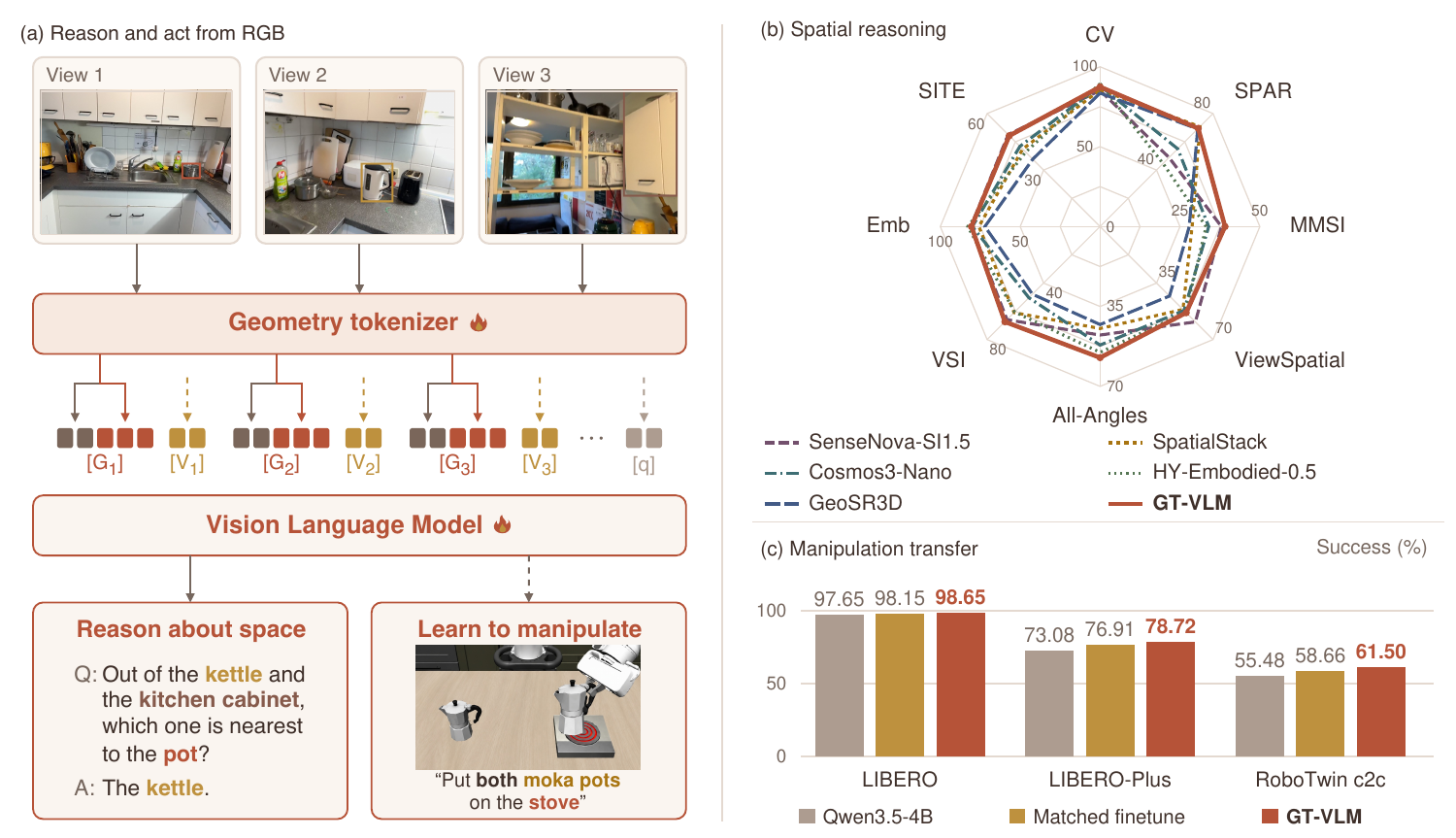}
\captionsetup{hypcap=false}
\captionof{figure}{\textbf{Learning to reason and act with geometry tokens.} GT-VLM combines geometry tokens ($G_i$) with vision tokens ($V_i$) for stronger spatial reasoning and GROOT policies. Illustrative tasks appear at left; the plots show spatial performance and manipulation success after the same GROOT post-training. The radar uses axis-specific scales; LIBERO averages four suites.}
\label{fig:overview}
\captionsetup{hypcap=true}
\end{strip}

\setcounter{dbltopnumber}{1}
\begin{abstract}
Vision-language models achieve strong semantic understanding but still struggle with spatial reasoning. Recent approaches address this gap by incorporating pretrained 3D geometric priors. We introduce GT-VLM, which organizes these priors to jointly expose local spatial structure and global scene context for reasoning. Our tokenizer integrates multi-level geometry through DPT refinement and a native-feature bypass to form spatially compressed local tokens, while separately adapting camera and register features into global tokens. Dedicated geometry inputs complement semantic vision through spatial alignment with RGB observations. We jointly train the tokenizer and language-model adapters through single-stage language supervision. GT-VLM demonstrates strong video and multi-view spatial reasoning, achieving the highest spatial and overall averages in our comparison: 62.87\% and 66.00\% across eight spatial and two general visual benchmarks. Its spatial average exceeds SenseNova-SI1.5 and GeoSR3D by 3.59 and 10.09 percentage points, respectively. Beyond question answering, the learned representation provides a stronger initialization for robotic manipulation. With a GROOT action head and the same benchmark-specific action post-training recipe, GT-VLM improves success over instruction-only finetuning by 0.50, 1.81, and 2.84 percentage points on LIBERO, LIBERO-Plus, and RoboTwin c2c, respectively.
\end{abstract}

\section{Introduction}
Vision-language models (VLMs) have strong semantic understanding, yet spatial benchmarks reveal persistent difficulties in estimating distances, interpreting object arrangements, and reasoning across viewpoints~\cite{vsi,blink,mmsi}. These tasks require the model to connect what it sees with the underlying 3D structure of the scene. This capability supports spatial question answering and robot actions that depend on object locations and relations.

Pretrained geometry encoders recover 3D structure from RGB observations~\cite{vggt,vggtomega}. For place recognition, UniPR-3D~\cite{deng2026unipr} aggregates patch and register tokens but omits the camera token to favor viewpoint invariance. For spatial questions, the answer can depend on the viewpoint itself. Recent VLMs fuse geometric and visual features~\cite{vgllm,spatialmllm}, combine spatial and camera tokens~\cite{vlm3r,spacemind}, or inject geometry at multiple decoder layers~\cite{spatialstack}. How should a VLM present local geometry together with the scene context needed to interpret it across views?

Spatial reasoning requires local geometric evidence to be interpreted within the wide scene. We therefore construct a representation that brings together information from multiple encoder depths at each local position, while preserving the encoder's compact global outputs through a separate adapter. Spatial compression then controls the size of the local feature grid. To combine this representation with semantic vision, we provide geometry and RGB as distinct inputs linked by spatial coordinates. This keeps both sources accessible to the language model as it processes a question, while connecting local geometric evidence to the corresponding image regions.

Following this design, we introduce \textbf{GT-VLM}, built on Qwen3.5-4B and a frozen VGGT-Omega encoder. Its tokenizer uses DPT refinement to integrate four geometric feature levels, with a native-feature bypass that combines the refined output with the encoder's final-stage patch representation. The resulting local features are spatially compressed to control the number of language-model inputs. A separate adapter maps camera/register features into global tokens. We jointly train the tokenizer and language-model adapters through single-stage language supervision.

Across eight spatial and two general visual benchmarks, GT-VLM achieves spatial and overall averages of 62.87\% and 66.00\%, exceeding SenseNova-SI1.5 by 3.59 and 2.96 percentage points. It also outperforms GeoSR3D, GeoVR-4B, and Spatial-MLLM-1.1 under the same evaluation inputs. The proposed interface achieves a higher spatial average than alternative geometry designs. Geometry intervention experiments further show that the model uses scene-specific content.

To test whether the representation is useful beyond question answering, we use the same instruction-tuned checkpoint to initialize policies with a GROOT action head. After post-training on robot demonstrations, GT-VLM improves success on LIBERO, LIBERO-Plus, and RoboTwin c2c over both the original Qwen3.5 backbone and the instruction-only finetuned model under the same action-learning recipe.

Our contributions are threefold.
\begin{itemize}
\item \textbf{A local-global tokenizer for 3D priors.}\nopagebreak[4] Our tokenizer organizes pretrained 3D priors into dedicated local and global inputs. DPT refinement and the native-feature bypass integrate features from multiple encoder depths into the local representation, while a separate adapter provides global scene information.
\item \textbf{Stronger spatial reasoning through structured geometry.}\nopagebreak[4] GT-VLM achieves an eight-benchmark spatial average of 62.87\%, exceeding SenseNova-SI1.5 and GeoSR3D by 3.59 and 10.09 percentage points, respectively. Interface ablations further support the proposed geometry representation and input design.
\item \textbf{A better initialization for action learning.}\nopagebreak[4] Used to initialize GROOT policies, the same instruction-tuned checkpoint improves success over instruction-only finetuning by 0.50, 1.81, and 2.84 percentage points on LIBERO, LIBERO-Plus, and RoboTwin c2c with the same action post-training recipe.
\end{itemize}

\section{Related Work}
\paragraph{Geometry-augmented vision-language models.}
VG-LLM~\cite{vgllm} and Spatial-MLLM~\cite{spatialmllm} fuse pretrained geometric and semantic visual features for RGB-based spatial reasoning. SpatialStack~\cite{spatialstack} aligns multi-level geometry with different language-decoder depths to capture local detail and global context, while GUIDE~\cite{guide} studies geometry injection within the decoder. GeoSR~\cite{geosr} encourages geometry use through visual masking and gated fusion, with a dynamic branch that also appends geometry tokens. Other approaches change the training objective or deployment architecture: G$^2$VLM~\cite{g2vlm} couples spatial understanding with geometric reconstruction, while GeoVR~\cite{geovr} transfers geometric supervision into a VLM and removes the external geometry encoder at deployment.

\paragraph{Geometry tokens and interface design.}
VLM-3R~\cite{vlm3r} fuses spatial and camera-view tokens with visual features through cross-attention; SpaceMind~\cite{spacemind} uses camera features to modulate spatial tokens. VLM-IE3D~\cite{vlmie3d} combines implicit and explicit geometry, while GeoReader~\cite{georeader} compares ways to read and inject continuous summaries. Learning to Reason in 4D~\cite{gsm} conditions its geometry inputs on the question. In driving scenes, GaussianDWM~\cite{Deng_2026_CVPR} selects 3D Gaussians for grounding based on the query; GaussianDWM++~\cite{deng2026gaussiandwm++} pools them into world tokens with text-conditioned cross-attention. GT-VLM instead constructs its geometry tokens from the observations alone. DPT refinement and a native-feature bypass supply local tokens, while a separate adapter preserves camera and register tokens alongside RGB.

VGGT-Occ~\cite{chen2026vggt} uses projection-aware attention to lift frozen VGGT features into a 3D occupancy grid. In language models, DK-VLM~\cite{dkvlm} uses geometry-dependent attention kernels, while SpaTime~\cite{spatime} studies streaming spatio-temporal reasoning.

\paragraph{Geometric representations for manipulation.}
VGGT-Omega~\cite{vggtomega} supplies frozen scene registers as additional OpenVLA-OFT inputs, while GeoAlign~\cite{geoalign} uses robot-state queries to select geometry for action prediction. SpatialBot~\cite{spatialbot} adapts spatially trained RGB-depth VLMs to manipulation using SpatialQA-E. SPEAR-1~\cite{spear1} trains a geometry-aware VLM on 3D question answering before adding an action expert trained on robot demonstrations.

\section{Geometry Tokens for Reasoning and Action}
\label{sec:method}

\subsection{Overview and Scene Correspondence}
We construct scene-aligned geometry inputs so that language reasoning can use local structure and global context alongside semantic vision. The tokenizer integrates features from different depths of the geometry encoder into a local representation before compression and adapts its global outputs separately. Geometry and RGB occupy distinct input positions, with shared spatial coordinates linking them to the same observations.

\begin{figure*}[t]
\centering
\includegraphics[width=\textwidth]{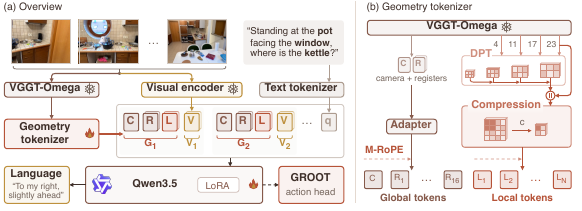}
\caption{\textbf{GT-VLM architecture.} Dedicated local and global geometry tokens complement semantic vision for spatial reasoning. (a) Each observation contributes geometry and RGB tokens to an interleaved sequence. The dashed GROOT branch denotes benchmark-specific action post-training. (b) DPT refinement and a native stage-23 bypass construct local features before compression; a separate adapter produces global tokens from camera/register features. M-RoPE aligns the geometry tokens with their corresponding RGB coordinates.}
\label{fig:architecture}
\end{figure*}

Let $\mathcal{I}=(I_1,\ldots,I_F)$ denote the RGB observations for a prompt $q$: a single image, ordered views, or video frames. A semantic encoder produces visual tokens $V_f$, and a frozen geometry encoder extracts multi-level features from the corresponding observations. A trainable tokenizer $T_\theta$ maps these features to geometry embeddings $G_f$. The language decoder receives geometry, vision, and text in a shared embedding space of dimension $d=2560$ (Fig.~\ref{fig:architecture}).

To reason jointly over $G_f$ and $V_f$, the two representations must refer to the same observations and image regions. The visual processor determines image resizing and video frame grouping, whereas the geometry encoder produces features on its own patch grid. Geometry inputs therefore follow the visual processor's resizing and temporal grouping, and their positions are mapped into the RGB coordinate system.

\subsection{Multi-scale Aggregation with a Native Bypass}
VGGT-Omega~\cite{vggtomega} supplies features from stages $\mathcal{S}=\{4,11,17,23\}$. For a scene canvas with patch grid $H\times W$, the patch features are $P_f^{(s)}\in\mathbb{R}^{H\times W\times2048}$, and the final-stage global features are $R_f\in\mathbb{R}^{17\times2048}$. The encoder combines within-frame and inter-frame information. Its global sequence carries cross-frame information through one camera token and sixteen register tokens.

To integrate features from the four encoder depths at each local position, a DPT-style pyramid~\cite{dpt} applies scale-specific projection and top-down refinement using trainable copies of the pretrained dense-head layers (Fig.~\ref{fig:architecture}b). We pool the refined output to the native geometry patch grid to obtain $D_f$. A parallel bypass provides the final-stage features $P_f^{(23)}$ directly. The two branches are fused as
\begin{equation}
 U_f = \phi\!\left([A_D(D_f)\,\Vert\,A_N(P_f^{(23)})]\right),
 \label{eq:fusion}
\end{equation}
where $A_D$ and $A_N$ project the DPT and native features to 512 channels each, $\Vert$ denotes channel concatenation, and $\phi$ is a learned $1\times1$ convolution. The fused grid $U_f\in\mathbb{R}^{H\times W\times1024}$ has one feature vector at each geometry-patch position.

\subsection{Local Compression and Global Adaptation}
\paragraph{Local geometry.}
The local compressor reduces the spatial resolution of $U_f$ to produce fewer language-model inputs while preserving the grid layout. It applies layer normalization, a depthwise strided convolution, and an MLP projection to dimension $d$. For compression factor $c$, the convolution uses kernel $c+1$, stride $c$, and padding $c/2$. Feature-grid padding handles dimensions that are not divisible by $c$.

The compressed grid has $H_c=\lceil H/c\rceil$ and $W_c=\lceil W/c\rceil$ positions. Flattening it in spatial order yields local embeddings $L_f\in\mathbb{R}^{H_cW_c\times d}$. We use $c=2$ for GT-VLM.

\paragraph{Global geometry.}
The camera/register sequence provides scene-level context alongside the local grid. The global adapter $A_R$ maps these features $R_f$ to seventeen $d$-dimensional embeddings per frame using layer normalization and a $2048\!\rightarrow\!1024\!\rightarrow\!2560$ MLP with GELU (Fig.~\ref{fig:architecture}b). Concatenating them with the local tokens gives $G_f=[A_R(R_f);L_f]$, where $[\,;\,]$ denotes token-axis concatenation.

Across $F$ observations, the geometry-token count is
\begin{equation}
 N_{\mathrm{geo}}=F\left(17+\left\lceil\frac{H}{c}\right\rceil
 \left\lceil\frac{W}{c}\right\rceil\right).
 \label{eq:budget}
\end{equation}
A $32\times32$ native patch grid yields 273, 81, and 33 geometry tokens per frame for $c=2,4,8$.

\subsection{Dedicated Input Slots and Aligned M-RoPE}
We place each geometry block $G_f$ immediately before its RGB block $V_f$, preserving separate positions for spatial and semantic evidence (Fig.~\ref{fig:architecture}a). Denote the full ordered context by $X$; a prompt with trailing text has $X=[G_1;V_1;\ldots;G_F;V_F;q]$.

VGGT requires observations from the same scene to have a common input size, so we pad mixed-resolution images to a shared canvas. The RGB branch retains their individual resized dimensions, giving corresponding image regions different coordinates in the two branches. We align the two branches through multimodal rotary position embeddings (M-RoPE)~\cite{qwen2vl,qwen35}, expressing geometry positions in RGB coordinates after accounting for the padding offsets and local compression stride. Each $(G_f,V_f)$ pair shares a frame identity and spatial origin, while camera/register tokens use the scene-grid center. Qwen3.5 tracks token order separately from this spatial correspondence.

\subsection{Single-stage Instruction Tuning}
\label{sec:training}
Starting from Qwen3.5-4B~\cite{qwen35}, we adapt the language model with rank-32 LoRA~\cite{lora} and jointly train the geometry tokenizer. The geometry encoder is frozen. Given the ordered context $X$ and assistant answer tokens $a_1,\ldots,a_T$, we minimize
\begin{equation}
 \mathcal{L}_{\mathrm{QA}}=-\sum_{t=1}^{T}\log p_{\theta,\psi}
 (a_t\mid X,a_{<t}),
\end{equation}
where $\theta$ and $\psi$ are the trainable tokenizer and LoRA parameters, respectively. Answer supervision trains the model to draw on geometric and semantic evidence when answering spatial questions. We optimize this language-only objective in a single stage for 72,000 optimizer steps with global batch size 96.

\begin{table*}[t]
\centering
\setlength{\tabcolsep}{1.8pt}
\small
\begin{tabular*}{\textwidth}{@{\extracolsep{\fill}}l*{13}{r}@{}}
\toprule
& & \multicolumn{9}{c}{Spatial Reasoning} & \multicolumn{2}{c}{General} & Overall \\
\cmidrule(lr){3-11}\cmidrule(lr){12-13}
Model & \shortstack{Params\\(B)} & CV & SPAR & MMSI & View & Angles & VSI & Emb & SITE & \textbf{Avg.} & MMB & BLINK & Avg. \\
\midrule
\multicolumn{14}{l}{\emph{Approx. 8B scale}} \\
MiMo-Embodied & 8.3 & 87.14 & 41.51 & 31.90 & 36.10 & 49.00 & 48.49 & 76.24 & 36.44 & 50.85 & 86.82 & 56.60 & 55.02 \\
GeoSR3D & 9.3 & 83.52 & \rankthird{69.11} & 27.90 & 43.03 & 42.92 & 47.89 & 72.12 & 35.77 & 52.78 & 85.56 & 51.76 & 55.96 \\
Embodied-R1.5 & 8.8 & 86.52 & 40.02 & 29.80 & 43.70 & 48.40 & 59.20 & 76.00 & 41.14 & 53.10 & \rankthird{87.94} & 57.29 & 57.00 \\
Cosmos3-Nano & 8.0 & \ranksecond{88.07} & 55.01 & \rankthird{34.00} & 52.00 & \rankthird{51.90} & 50.40 & 80.00 & \rankthird{42.59} & 56.75 & \ranksecond{88.30} & 59.13 & 60.14 \\
SenseNova-SI1.5 & 7.9 & 85.18 & 48.85 & \ranksecond{38.00} & \rankfirst{59.05} & 47.51 & \ranksecond{65.92} & \ranksecond{80.91} & \rankfirst{48.83} & \ranksecond{59.28} & 86.84 & \rankfirst{69.33} & \ranksecond{63.04} \\
\midrule
\multicolumn{14}{l}{\emph{Approx. 4B scale}} \\
GeoVR-4B & 4.4 & 63.41 & 32.19 & 23.90 & 35.07 & 34.99 & 53.62 & 37.42 & 16.92 & 37.19 & 61.50 & 40.35 & 39.94 \\
Spatial-MLLM-1.1 & 5.0 & 72.73 & 34.17 & 25.20 & 39.37 & 41.93 & 53.28 & 64.84 & 26.05 & 44.70 & 81.29 & 43.08 & 48.19 \\
RoboBrain 2.5-4B & 4.4 & \rankthird{87.66} & 43.53 & 20.50 & 36.60 & 43.80 & 41.70 & 73.80 & 36.48 & 48.01 & 87.37 & 67.33 & 53.88 \\
Qwen3.5-4B & 4.5 & 70.58 & 39.16 & 30.10 & 43.79 & 51.27 & 53.60 & 72.14 & 35.31 & 49.49 & 80.79 & 62.49 & 53.92 \\
HY-Embodied-0.5 & 3.8 & \rankfirst{89.39} & 44.62 & 33.20 & \rankthird{53.10} & \ranksecond{55.10} & 60.50 & \rankfirst{82.80} & 39.55 & 57.28 & 84.19 & \rankthird{67.60} & \rankthird{61.00} \\
SpatialStack & 5.6 & 85.53 & \rankfirst{70.71} & 28.80 & 51.33 & 44.65 & \rankthird{60.90} & 75.91 & 41.12 & \rankthird{57.37} & 86.91 & 57.50 & 60.34 \\
\midrule
\textbf{GT-VLM (c2)} & 5.5 & 87.40 & \ranksecond{69.51} & \rankfirst{39.10} & \ranksecond{53.36} & \rankfirst{57.41} & \rankfirst{67.42} & \rankthird{80.52} & \ranksecond{48.23} & \rankfirst{62.87} & \rankfirst{88.51} & \ranksecond{68.54} & \rankfirst{66.00} \\
\bottomrule
\end{tabular*}

\caption{\textbf{Performance on spatial and general visual benchmarks (\%).} GT-VLM achieves the highest spatial and overall averages. Models are grouped by approximate parameter scale. Spatial Avg. and Overall Avg. are the equal-weight means over the eight spatial benchmarks and all ten benchmarks, respectively. Shading indicates \ranklegend\ in each column.}
\label{tab:vqa}
\end{table*}

To develop spatial reasoning alongside broader instruction following, we combine 7,999,872 examples from ten sources and 73 components, organized into spatial, general visual, text, and embodied-language groups with an approximate 80:10:5:5 ratio. The 72K training run draws 6,912,000 examples: 5,529,216 spatial, 691,968 general visual, 345,472 text, and 345,344 embodied-language examples. The spatial pool combines geometric reasoning, counting, grounding, and mixed QA from SPAR~\cite{spar}, CA-VQA~\cite{cavqa}, SenseNova-SI-8M~\cite{sensenovasi}, JoyAI/OpenSpatial~\cite{openspatial}, and VLM-3R~\cite{vlm3r}. The remaining groups use LLaVA visual/text instruction~\cite{onevision}, Cosmos/RoboVQA~\cite{cosmosreason1,robovqa}, EgoIT~\cite{egolife}, and RoboInter~\cite{robointer}.

\subsection{Action Post-training}
For manipulation, we use the instruction-tuned VLM to initialize a policy with a GROOT action head~\cite{groot}. We then post-train the policy on robot demonstrations for each benchmark, learning action predictions conditioned on the visual observations and language instruction.

\section{Experiments}
\subsection{Evaluation Setup}
\begin{figure}[t]
\centering
\includegraphics[width=\linewidth]{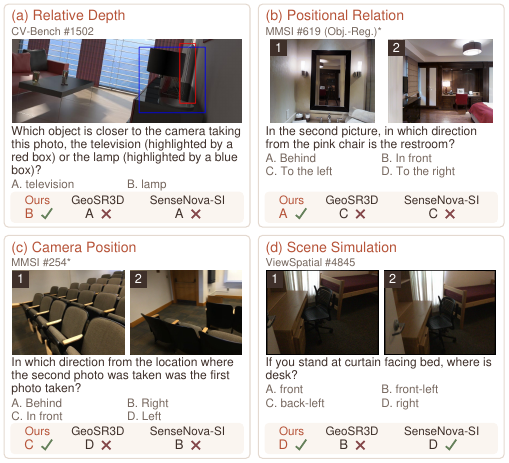}
\caption{\textbf{Selected spatial reasoning examples.} GT-VLM answers the selected depth, relation, and viewpoint questions correctly. Check/cross marks indicate correct/incorrect answers; * marks the dataset's hard examples.}
\label{fig:spatialqa}
\end{figure}

\begin{figure*}[t]
\centering
\includegraphics[width=\textwidth]{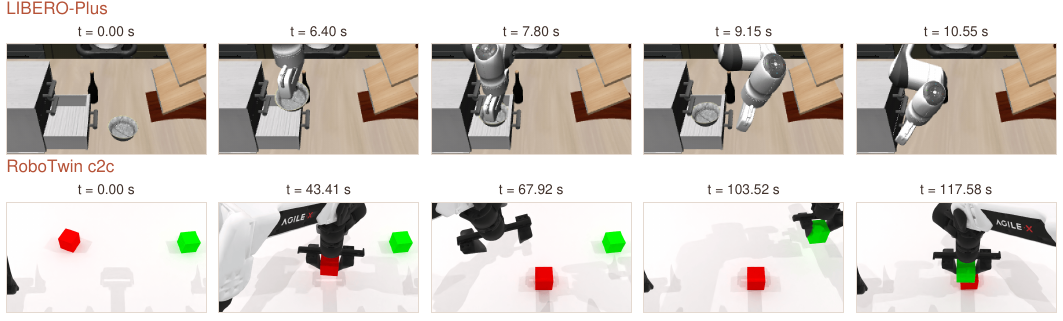}
\caption{\textbf{Simulation rollouts.} The sequences show multi-step object manipulation: bowl placement and drawer closing in LIBERO-Plus, and successful GT-VLM block stacking in RoboTwin c2c.}
\label{fig:rollouts}
\end{figure*}

\paragraph{Compared models.}
We compare GT-VLM c2 with Qwen3.5~\cite{qwen35}, MiMo-Embodied~\cite{mimoembodied}, Embodied-R1.5~\cite{embodiedr15}, Cosmos3-Nano~\cite{cosmos3}, SenseNova-SI1.5~\cite{sensenovasi}, RoboBrain 2.5~\cite{robobrain25}, HY-Embodied-0.5~\cite{hyembodied05}, and SpatialStack~\cite{spatialstack} (Table~\ref{tab:vqa}). GeoSR3D~\cite{geosr}, GeoVR-4B~\cite{geovr}, and Spatial-MLLM-1.1~\cite{spatialmllm} use the same evaluation inputs as c2.

\paragraph{Benchmarks and inputs.}
We evaluate spatial reasoning on CV-Bench~\cite{cvbench}, SPAR-Bench~\cite{spar}, MMSI-Bench~\cite{mmsi}, ViewSpatial-Bench~\cite{viewspatial}, All-Angles-Bench~\cite{allangles}, VSI-Bench~\cite{vsi}, EmbSpatial-Bench~\cite{embspatial}, and SITE~\cite{site}. The development split of MMBench~\cite{mmbench} and the BLINK validation set~\cite{blink} assess broader visual understanding. For VSI-Bench, we uniformly sample 32 frames per video and provide them as an ordered image sequence. CV-Bench, MMBench-Dev, and EmbSpatial use single-image inputs; the other image benchmarks use their released single-view or multi-view observations.

Spatial Avg. is the equal-weight mean of the eight spatial benchmarks; Overall Avg. averages all ten benchmarks.

\subsection{Spatial Reasoning and General Vision}
\label{sec:vqa}
GT-VLM achieves the highest Spatial Avg. (62.87) and Overall Avg. (66.00) in Table~\ref{tab:vqa}, improving over SenseNova-SI1.5 by 3.59 and 2.96 points. Its spatial average also exceeds SpatialStack, HY-Embodied-0.5, and Cosmos3-Nano by 5.50, 5.59, and 6.12 points. The model leads on VSI (67.42), MMSI (39.10), and All-Angles (57.41), showing strong performance when spatial judgments involve video or multiple views of a scene.

The input-aligned comparisons evaluate GT-VLM against other geometry-based VLMs on the same observations. GT-VLM outperforms GeoSR3D, GeoVR-4B, and Spatial-MLLM-1.1 on all ten benchmarks, with spatial-average gains of 10.09, 25.68, and 18.17 points. GeoSR3D is the strongest of these three baselines in spatial average. Against it, the gains reach 19.53 points on VSI and 14.49 on All-Angles, highlighting the advantage on video and cross-view reasoning.

Figure~\ref{fig:spatialqa} presents selected examples of relative depth, object-region relations, camera position, and observer-relative directions, comparing GT-VLM with GeoSR3D and SenseNova-SI1.5.

Combining geometry with semantic vision also yields strong results beyond the spatial benchmarks: GT-VLM ranks first on MMBench-Dev (88.51) and second on BLINK (68.54).

\setcounter{topnumber}{1}
\setlength{\textfloatsep}{12pt plus 2pt minus 2pt}
\setlength{\dbltextfloatsep}{12pt plus 2pt minus 2pt}
\begin{table}[t]
\centering\small
\renewcommand{\arraystretch}{1.23}
\setlength{\tabcolsep}{2.5pt}
\begin{tabular*}{\linewidth}{@{\extracolsep{\fill}}lrrr@{}}
\toprule
Method & LIBERO & \shortstack{LIBERO-\\Plus} & \shortstack{RoboTwin\\c2c} \\
\midrule
\multicolumn{4}{l}{\emph{References}} \\
OpenVLA~\cite{openvla} & 76.50 & 15.60 & -- \\
OpenVLA-OFT~\cite{openvlaoft} & 97.10 & 69.60 & -- \\
$\pi_0$~\cite{pi0} & 94.15 & 53.60 & 46.42 \\
$\pi_{0.5}$~\cite{pi05} & 96.85 & -- & 70.66 \\
X-VLA~\cite{xvla} & 98.10 & -- & 67.96 \\
Xiaomi Robotics-0~\cite{xiaomirobotics0} & 98.70 & -- & 62.86 \\
\midrule
\multicolumn{4}{l}{\emph{Matched GROOT controls}} \\
Base + GROOT & 97.65 & 73.08 & 55.48 \\
Finetune + GROOT & 98.15 & 76.91 & 58.66 \\
\textcolor{rankwarm}{GT-VLM + GROOT} & \textbf{98.65} & \textbf{78.72} & \textbf{61.50} \\
\bottomrule
\end{tabular*}

\caption{\textbf{Manipulation success (\%).} GT-VLM improves on both matched GROOT controls across all three benchmarks.}
\label{tab:vla}
\end{table}
\begin{table}[t]
\centering\small
\renewcommand{\arraystretch}{0.90}
\setlength{\tabcolsep}{2.5pt}
\begin{tabular*}{\linewidth}{@{\extracolsep{\fill}}lrrr@{}}
\toprule
Suite / perturbation & Base & Finetune & \textcolor{rankwarm}{GT-VLM} \\
\midrule
\multicolumn{4}{l}{\emph{LIBERO suites}} \\
Spatial & 98.00 & 98.40 & \textbf{98.60} \\
Object & 99.20 & 99.20 & \textbf{99.40} \\
Goal & \textbf{99.60} & 99.40 & 99.20 \\
LIBERO-10 & 93.80 & 95.60 & \textbf{97.40} \\
\cmidrule(lr){1-4}
Avg. & 97.65 & 98.15 & \textbf{98.65} \\
\midrule
\multicolumn{4}{l}{\emph{LIBERO-Plus perturbations}} \\
Camera view & 41.65 & 56.97 & \textbf{63.91} \\
Sensor noise & 78.33 & \textbf{86.38} & 84.63 \\
Object layout & 80.98 & 81.44 & \textbf{82.95} \\
Robot initial state & 49.16 & \textbf{52.77} & 51.87 \\
Lighting & 95.36 & 94.75 & \textbf{95.88} \\
Background & \textbf{95.17} & 93.12 & 94.24 \\
Language & 84.58 & 83.02 & \textbf{87.25} \\
\cmidrule(lr){1-4}
Avg. & 73.08 & 76.91 & \textbf{78.72} \\
\bottomrule
\end{tabular*}

\caption{\textbf{Matched-control breakdown (\%).} GT-VLM gains most over finetuning on LIBERO-10 and camera-view changes.}
\label{tab:perturbations}
\end{table}

\subsection{Transfer to Robotic Manipulation}
\label{sec:vla}
To test whether the representation learned for question answering provides a better starting point for action learning, we compare GROOT policies initialized from original Qwen3.5, Qwen3.5 finetuned on our instruction data, and GT-VLM c2. All three use the same benchmark-specific action post-training recipe. LIBERO~\cite{libero} equally weights four suites, each with ten tasks and fifty episodes per task, totaling 2,000 evaluation episodes. LIBERO-Plus~\cite{liberoplus} weights its seven perturbation categories by episode count. RoboTwin c2c~\cite{robotwin,robotwin2} (Fig.~\ref{fig:rollouts}) uses 2,500 clean demonstration episodes for 30K post-training steps, followed by evaluation on fifty tasks with one hundred clean episodes each.

GT-VLM achieves 98.65\% on LIBERO, 78.72\% on LIBERO-Plus, and 61.50\% on RoboTwin c2c (Table~\ref{tab:vla}). Gains over original Qwen3.5 are 1.00, 5.64, and 6.02 points; gains over matched finetuning are 0.50, 1.81, and 2.84 points. The improvement over the instruction-only control shows that the geometry-augmented VLM provides a more effective starting point for action learning.

LIBERO performance is already close to saturation on the Spatial, Object, and Goal suites, with all three policies at or above 98\%. The main gain comes from LIBERO-10, where GT-VLM reaches 97.40\% compared with 95.60\% for the finetuned control and 93.80\% for the original backbone. Figure~\ref{fig:rollouts} shows rollouts on LIBERO-Plus and RoboTwin c2c.

\begin{figure*}[t]
\centering
\includegraphics[width=\textwidth]{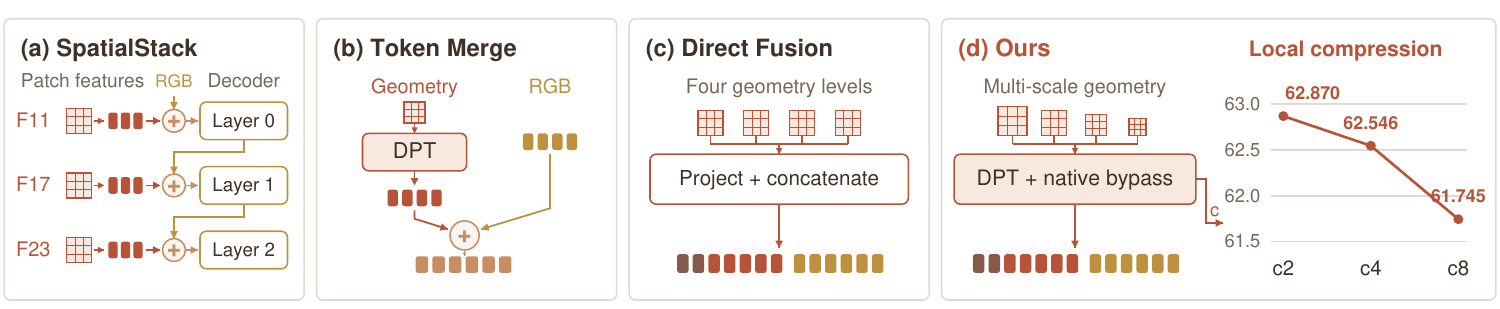}
\caption{\textbf{Geometry interfaces and local compression.} GT-VLM refines multi-level geometry before local compression. SpatialStack~\cite{spatialstack} is a decoder-injection reference; Token Merge adds DPT-derived geometry to visual tokens, while Direct Fusion and GT-VLM use dedicated inputs. The c2/c4/c8 curve shows decreasing eight-benchmark Spatial Avg. with stronger local compression.}
\label{fig:interface}
\end{figure*}

LIBERO-Plus examines whether the policies remain effective as their observations and instructions change. The largest gain over finetuning occurs under camera-view changes (Table~\ref{tab:perturbations}). Success rises from 41.65\% for original Qwen3.5 and 56.97\% for finetuning to 63.91\% for GT-VLM, a 6.94-point gain over the finetuned control. Language-perturbation success improves by 4.23 points.

\section{Analysis and Ablations}

\subsection{Geometry-token Design}
\label{sec:interface}
Our tokenizer builds a local/global geometry representation before presenting it alongside semantic vision. We compare this organization with instruction-only training and alternative geometry interfaces (Table~\ref{tab:reasoningablation}). Finetuning Qwen3.5 on the same instruction data yields a Spatial Avg. of 59.62, compared with 62.87 for GT-VLM.

\begin{table}[t]
\centering\small
\renewcommand{\arraystretch}{0.98}
\setlength{\tabcolsep}{2.5pt}
\begin{tabular*}{\linewidth}{@{\extracolsep{\fill}}lrrr@{}}
\toprule
Variant & \shortstack{Spatial\\Avg.} & VSI & Angles \\
\midrule
Qwen3.5 finetune & 59.62 & 63.34 & 51.97 \\
SpatialStack-style & 61.84 & 66.34 & 53.28 \\
Direct Fusion & 61.84 & 66.53 & 52.20 \\
Token Merge & 61.61 & 66.61 & 51.69 \\
PatchOnly & 61.97 & 66.90 & 53.19 \\
\midrule
\textcolor{rankwarm}{GT-VLM c2} & \textbf{62.87} & \textbf{67.42} & \textbf{57.41} \\
\bottomrule
\end{tabular*}

\caption{\textbf{Spatial reasoning ablations (\%).} GT-VLM leads in Spatial Avg., with the largest spatial gains on All-Angles. Avg.: eight spatial benchmarks; Angles: All-Angles.}
\label{tab:reasoningablation}
\end{table}
\begin{table}[t]
\centering\small
\renewcommand{\arraystretch}{0.96}
\setlength{\tabcolsep}{1.5pt}
\begin{tabular*}{\linewidth}{@{\extracolsep{\fill}}lrrl@{}}
\toprule
Intervention & VSI & $\Delta$ (pp) & 95\% paired CI \\
\midrule
Scene-matched & 67.42 & 0.00 & N/A \\
All / wrong scene & 63.52 & -3.90 & [-5.08, -2.69] \\
All / zero & 58.54 & -8.88 & [-10.53, -7.30] \\
\midrule
Patches / wrong scene & 66.60 & -0.82 & [-1.33, -0.23] \\
Patches / zero & 61.11 & -6.31 & [-7.54, -4.89] \\
C/R / wrong scene & 63.28 & -4.14 & [-5.21, -2.88] \\
C/R / zero & 63.29 & -4.13 & [-5.33, -2.85] \\
\bottomrule
\end{tabular*}

\caption{\textbf{Geometry interventions on VSI.} Mismatched and zeroed geometry both lower accuracy relative to scene-matched inputs. C/R: camera/register tokens.}
\label{tab:intervention}
\end{table}

\paragraph{Geometry input organization.}
The SpatialStack-style variant offers an alternative use of multi-level geometry: features from encoder layers 11, 17, and 23 are processed separately and added at decoder layers 0, 1, and 2 (Fig.~\ref{fig:interface}). With the same training data and tokenizer learning rate as GT-VLM, it reaches a Spatial Avg. of 61.84. GT-VLM's 1.03-point gain favors constructing a local/global representation before combining it with semantic vision.

\paragraph{Local feature construction.}
To build each local position from multiple encoder depths, GT-VLM uses DPT refinement and a native-feature bypass. Both Direct Fusion and GT-VLM use four geometry feature levels. Direct Fusion combines them by projection and concatenation; our construction additionally refines information across scales while maintaining direct access to the final-stage features. With the same single-stage 72K training, data, and learning-rate schedule, GT-VLM improves Spatial Avg. by 1.03 points over Direct Fusion.

\paragraph{Local and global geometry.}
Scene-level camera/register features complement the region-specific evidence in local tokens. PatchOnly is a local-only variant that omits the global tokens from the input sequence. Compared with this variant, GT-VLM gains 0.90 points in Spatial Avg., supporting the inclusion of scene-level context alongside local geometry.

\paragraph{Dedicated geometry inputs.}
The final choice is how the constructed geometry meets semantic vision. Token Merge constructs geometry tokens with DPT and then adds them to RGB features at the visual-token positions. GT-VLM keeps the sources at separate positions, giving the decoder distinct attention targets linked by spatial correspondence. GT-VLM improves Spatial Avg. by 1.26 points over Token Merge. Because Token Merge also uses DPT, this result highlights the role of input organization alongside geometric feature construction.

Across all four interface variants, the largest gains among the eight spatial benchmarks occur on All-Angles, ranging from 4.13 to 5.72 points. This concentration of gains on cross-view reasoning matches the goal of providing spatially organized local evidence together with scene context.

\subsection{Scene-specific Geometry}
\label{sec:intervention}
To examine whether the model uses the scene content exposed by this interface, we replace its adapted geometry features with those from another scene or set them to zero. The interventions act after input assembly, keeping RGB, text, positions, and sequence length fixed. All conditions use the same 5,130 VSI questions and 288 recipient scenes, with paired 95\% confidence intervals (Table~\ref{tab:intervention}).

Wrong-scene geometry lowers VSI from 67.42 to 63.52. The input still contains a populated geometry sequence, but the correspondence between geometry and the observed scene is broken. The drop shows that GT-VLM uses scene-specific geometric information. Zeroing all geometry causes a larger decrease to 58.54, removing that information while introducing a stronger distribution shift than substitution.

Perturbing the two branches separately reveals different responses to replacement and removal. Replacing local patches reduces performance by 0.82 points, whereas zeroing them causes a 6.31-point drop. Camera/register replacement and zeroing produce similar decreases of 4.14 and 4.13 points. The global branch is sensitive to mismatched scene content, while local features show a larger response to removal than replacement.

\begin{samepage}
\subsection{Compression and Inference Cost}
\label{sec:cost}
Dedicated geometry inputs add to the decoder's sequence length, motivating compression of the dense local grid. Increasing the stride from c2 to c4 and c8 changes Spatial Avg. from 62.870 to 62.546 and 61.745 (Fig.~\ref{fig:interface}). The reductions of 0.324 and 1.125 points are modest in aggregate, but All-Angles falls from 57.41 to 54.13 and 52.49. c8 scores 66.42 on VSI, approximately one point below c2. The larger losses on All-Angles motivate keeping c2 as the default and using stronger compression when a shorter geometry sequence is needed.

\end{samepage}
\begin{table}[b]
\centering\small
\setlength{\tabcolsep}{1pt}
\begin{tabular*}{\linewidth}{@{\extracolsep{\fill}}lrrrrrr@{}}
\toprule
& \multicolumn{2}{c}{Base} & \multicolumn{2}{c}{c2} & \multicolumn{2}{c}{c4} \\
\cmidrule(lr){2-3}\cmidrule(lr){4-5}\cmidrule(lr){6-7}
Set & Prefill & Total & Prefill & Total & Prefill & Total \\
\midrule
CV & 42.86 & 107.48 & 95.24 & 121.16 & 86.71 & 112.55 \\
Angles & 38.63 & 2320.41 & 76.10 & 85.12 & 60.48 & 120.46 \\
VSI & 146.84 & 1134.17 & 405.69 & 435.34 & 341.36 & 376.10 \\
\bottomrule
\end{tabular*}

\caption{\textbf{RTX 5090 answer latency (ms, P50).} Stronger compression lowers prefill time, while answer length affects total latency. Prefill: to first logits; Total: complete answer. Angles: All-Angles.}
\label{tab:cost}
\end{table}
\begin{figure}[t]
\centering
\includegraphics[width=\linewidth]{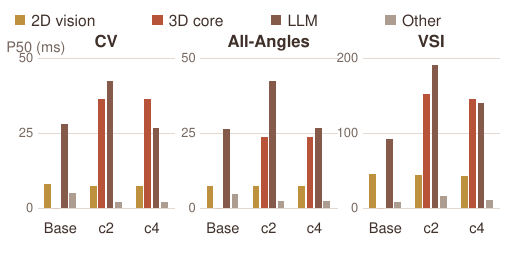}
\caption{\textbf{Prefill components on RTX 5090.} c4 chiefly reduces LLM prefill time. Bars show independent P50s; 3D core combines VGGT, DPT, and camera/register adaptation.}
\label{fig:prefill}
\end{figure}

We profile batch-one inference on a single RTX 5090 using BF16, merged LoRA, and KV caching. Table~\ref{tab:cost} reports prefill and complete-answer latency, and Figure~\ref{fig:prefill} breaks down prefill by component. Timing begins after media processing, device transfer, and graph preparation.

On VSI, c4 reduces total prefill from 405.69 to 341.36\,ms relative to c2. Complete-answer latency also depends on output length: on All-Angles, c2 produces shorter answers than c4 and finishes sooner despite slower prefill.

\subsection{Limitations}
Our experiments use one semantic backbone and one geometry encoder, so generalization across model families and scales remains to be tested. Geometry encoding and the additional decoder inputs increase inference cost, while local compression offers a tradeoff between input length and spatial accuracy. Manipulation transfer is currently evaluated in simulation. Real-robot experiments are needed to assess the learned representation under physical sensing and control conditions.

\section{Conclusion}
We presented GT-VLM, which organizes pretrained 3D priors into a local-global geometry representation for reasoning alongside semantic vision. Its tokenizer combines refined local features with separately adapted global scene information, preserving both as dedicated geometry inputs. This design achieves the highest spatial and overall averages in our comparison, with interface ablations supporting the value of its feature construction and input organization. Beyond question answering, the same instruction-tuned checkpoint provides a stronger initialization for GROOT policies and improves manipulation performance under matched action post-training.

\section*{LLM Usage Statement}
OpenAI Codex assisted with organizing related work, writing LaTeX and table/figure code, checking consistency, and editing experimental code. All numerical results come from the authors' experimental materials. The authors are responsible for the claims, citations, and final submission.

\clearpage

{\small\bibliographystyle{ieeenat_fullname}\bibliography{references}}

\clearpage
\addtolength{\topmargin}{-10pt}
\addtolength{\textheight}{27pt}
\onecolumn
\raggedbottom
\setlength{\intextsep}{9pt plus 1pt minus 1pt}
\appendix
\setcounter{table}{0}
\renewcommand{\thetable}{S\arabic{table}}
\renewcommand{\theHtable}{supp.\arabic{table}}
\enlargethispage{12pt}
{\centering\Large\bfseries Supplementary Material\par}
\section{Architecture and Training Details}
\subsection{Feature and Token Shapes}
The geometry encoder has patch size 16 and embedding dimension 1024. Its cached features concatenate frame-level and inter-frame representations, giving 2048-dimensional features at stages 4, 11, 17, and 23. The stage-23 token order is one camera token, sixteen register tokens, then dense patches. The DPT pathway uses independently trainable copies of the pretrained dense-head normalization, projection, resizing, and refinement modules. After refinement, features are pooled to the native geometry grid. Both the DPT and native-feature branches project to 512 dimensions; their concatenation is fused into 1024 channels.

To reduce the dense feature grid before language-model processing, the local compressor applies layer normalization, bottom/right feature padding to a multiple of $c$, and a depthwise convolution with kernel size $c+1$, stride $c$, and padding $c/2$. A linear embedding of the valid-feature fraction is added to the compressed features, followed by a $1024\rightarrow1024\rightarrow2560$ MLP with GELU. The separate camera/register adapter maps the seventeen global tokens through layer normalization and a $2048\rightarrow1024\rightarrow2560$ MLP with GELU.

\subsection{M-RoPE Coordinate Assignment}
Geometry inputs use the resized observations selected by the visual processor. For video, each geometry frame corresponds to one visual temporal patch group and uses its first actual frame. Mixed-resolution views are centered on a common white canvas for geometric encoding. Each geometry/RGB pair shares a positional origin and frame identity, with temporal, height, and width coordinates assigned to both blocks. RGB positions follow the backbone's native mapping.

Local geometry coordinates are expressed in RGB-grid units. With compression factor $c$ and visual spatial-merge factor $m$, moving one cell in the compressed geometry grid changes the corresponding height or width coordinate by $c/m$. We express the top and left canvas-padding offsets in these units and subtract them from the geometry coordinates. The coordinates remain continuous, including fractional values and positions outside the native visual grid. Each frame's camera token and sixteen registers share the scene-grid center coordinates.

For consecutive geometry/RGB pairs, the next shared positional origin is set beyond the maximum position reached by the previous pair. This assigns successive pairs nonoverlapping coordinate ranges. Qwen3.5's position-ID tensor also carries a strictly increasing token-order coordinate, which follows the assembled sequence independently of the shared temporal and spatial positions. Text keeps its original order relative to the media. The geometry placeholder reserves a sequence slot for an adapted continuous embedding and is suppressed during generation.

\subsection{Instruction-tuning Configuration}
All variants in Table~\ref{tab:fullablation} use the same training examples, example order, 72K-step budget, and optimization hyperparameters. Their geometry interfaces and compression factors follow the variant definitions. The Qwen3.5 instruction-only control follows the same example sequence. Each instruction-tuning run uses a single seed.

Both encoders use the same observations. Training images and frames are resized to between 4,096 and 307,200 pixels. By default, videos supply 2 to 128 frames at 2\,fps, subject to the media policy and sequence budget.

\begin{table}[H]
\centering\small
\begin{tabular*}{\linewidth}{@{\extracolsep{\fill}}lrlr@{}}
\toprule
\multicolumn{2}{c}{Model and batch configuration} & \multicolumn{2}{c}{Optimization} \\
\cmidrule(lr){1-2}\cmidrule(l){3-4}
Semantic VLM & Qwen3.5-4B & Optimizer & AdamW \\
Geometry encoder & Frozen VGGT-Omega & Optimizer steps & 72,000 \\
Main local compression & $c=2$ & Warmup steps & 500 \\
Maximum sequence length & 8,192 & Learning-rate scheduler & Cosine \\
Global batch size & 96 & Scheduler horizon & 83,332 steps \\
Ranks / per-rank microbatch & 16 / 1 & Adam $\beta_1,\beta_2$ & 0.9, 0.95 \\
Gradient accumulation & 6 & Adam $\epsilon$ / weight decay & $10^{-8}$ / 0.01 \\
Packing / flattening & Disabled / disabled & Gradient clipping & 1.0 \\
LoRA rank / alpha / dropout & 32 / 64 / 0.05 & Learning rates (tokenizer / LoRA) & $10^{-4}$ / $10^{-5}$ \\
\bottomrule
\end{tabular*}
\caption{\textbf{Single-stage instruction-tuning configuration.} The tokenizer and LoRA adapters minimize answer-token cross-entropy. The scheduler horizon corresponds to one finite-mixture epoch.}
\label{tab:supconfig}
\end{table}

\clearpage
\section{Instruction Data}
\subsection{Training Mixture}
We sampled without replacement within each of the 73 components to build a mixture of 7,999,872 examples from ten sources (Table~\ref{tab:sources}). We fixed the data order using blocks of 256 and ordering seed 3047. Over 72,000 steps with a global batch size of 96, the model saw 6,912,000 examples. About 80\% came from the spatial instruction pool, 10\% from general visual instruction, and 5\% each from text-only and embodied-language data (Fig.~\ref{fig:training}). Different questions about the same image count as separate examples.

\supppair{%
\begingroup\captionsetup{type=table}
\centering\small
\setlength{\tabcolsep}{2pt}\begin{tabular}{lrrr}
\toprule
Source & Selected & Used at 72K & Share (\%) \\
\midrule
SPAR & 1,935,173 & 1,672,517 & 24.20 \\
CA-VQA & 1,935,156 & 1,672,244 & 24.19 \\
SenseNova-SI-8M & 1,547,959 & 1,337,015 & 19.34 \\
JoyAI / OpenSpatial & 774,059 & 668,843 & 9.68 \\
VLM-3R training QA & 207,525 & 178,597 & 2.58 \\
LLaVA-OneVision visual & 800,000 & 691,968 & 10.01 \\
LLaVA text instruction & 400,000 & 345,472 & 5.00 \\
Cosmos / RoboVQA & 66,667 & 57,600 & 0.83 \\
EgoIT & 33,333 & 28,320 & 0.41 \\
RoboInter & 300,000 & 259,424 & 3.75 \\
\midrule
Total & 7,999,872 & 6,912,000 & 100.00 \\
\bottomrule
\end{tabular}

\caption{\textbf{Training-data sources.} Shares are computed over the 6,912,000 examples used in training.}
\label{tab:sources}
\par\endgroup
}{%
\begingroup\captionsetup{type=figure}\centering
\includegraphics[width=\linewidth]{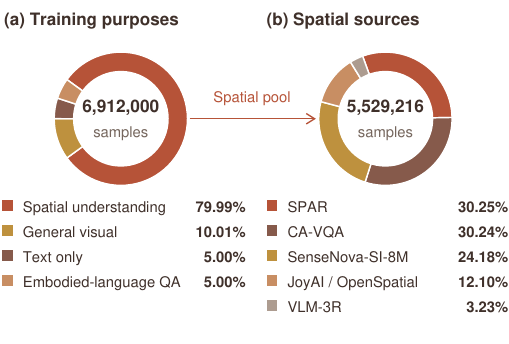}
\caption{\textbf{Instruction-tuning composition.} Left: training-purpose groups over 6,912,000 samples. Right: the 5,529,216 spatial-group samples by source.}
\label{fig:training}\par\endgroup

}
\supppair{%

The 1,672,517 SPAR examples used in instruction tuning come from the official SPAR-7M training split~\cite{spar}. The other spatial sources are CA-VQA from MM-Spatial~\cite{cavqa}, SenseNova-SI-8M~\cite{sensenovasi}, JoyAI-Image-OpenSpatial~\cite{openspatial}, and VLM-3R~\cite{vlm3r}. LLaVA-OneVision-Data~\cite{onevision} supplies the visual and text-only subsets; the text-only components include Evol-Instruct~\cite{evolinstruct}, Magpie~\cite{magpie}, and MathQA~\cite{mathqa}. The embodied-language sources are Cosmos-Reason1/RoboVQA~\cite{cosmosreason1,robovqa}, EgoIT-99K from EgoLife~\cite{egolife}, and RoboInter~\cite{robointer}.

\subsection{Task Composition}

Spatial-understanding data cover temporal order, camera/viewpoint correspondence, depth and distance, size, counting, 2D/3D grounding, navigation QA, and spatial relations. Table~\ref{tab:datatasks} gives the task composition, including mixed question-answer corpora and groups defined by answer format.

The embodied-understanding group contains language-supervised robot-video QA, egocentric instruction, contact and affordance, planning, scene understanding, success judgment, and trajectory questions. These examples provide language supervision. Robot demonstrations enter the training procedure when adapting the model for manipulation.
}{%
\begingroup\captionsetup{type=table}
\centering\small
\renewcommand{\arraystretch}{0.96}\setlength{\tabcolsep}{2pt}\begin{tabular}{lrr}
\toprule
Training task group & Sample positions & Share (\%) \\
\midrule
Temporal order & 48,640 & 0.70 \\
Camera/viewpoint correspondence & 147,408 & 2.13 \\
Geometry, depth, distance, size & 782,144 & 11.32 \\
Navigation QA & 10,205 & 0.15 \\
Counting/localization & 97,768 & 1.41 \\
Spatial relations/imagination & 586,352 & 8.48 \\
Spatial QA: answer-format labels & 836,506 & 12.10 \\
Counting/cardinality QA & 278,750 & 4.03 \\
2D grounding QA & 278,238 & 4.03 \\
3D grounding QA & 278,750 & 4.03 \\
Mixed spatial pool & 2,184,455 & 31.60 \\
Mixed general visual & 691,968 & 10.01 \\
Text instruction/reasoning & 335,744 & 4.86 \\
Text mathematics QA & 9,728 & 0.14 \\
Robot-video understanding QA & 57,600 & 0.83 \\
Egocentric image instruction & 7,160 & 0.10 \\
Egocentric video instruction & 21,160 & 0.31 \\
Contact/affordance & 54,770 & 0.79 \\
Temporal action understanding QA & 86,561 & 1.25 \\
Grounding QA & 13,479 & 0.20 \\
Planning QA & 40,977 & 0.59 \\
Scene understanding QA & 13,960 & 0.20 \\
Success judgment QA & 27,408 & 0.40 \\
Trajectory QA & 22,269 & 0.32 \\
\midrule
Total & 6,912,000 & 100.00 \\
\bottomrule
\end{tabular}

\caption{\textbf{Task-group exposure.} Percentages give each group's share of the 6,912,000 training examples.}
\label{tab:datatasks}
\par\endgroup
}

\clearpage
\section{Benchmark Protocols and Metric Definitions}
\subsection{Compared Models}
Parameter counts cover the components used for visual understanding, including encoders, adapters, and all inference experts. We count shared weights once, merge LoRA weights for inference, and report sizes to the nearest 0.1B. For Cosmos3-Nano~\cite{cosmos3}, we count the 8B understanding component.

We compare GT-VLM with Qwen3.5-4B~\cite{qwen35}, SenseNova-SI1.5~\cite{sensenovasi}, SpatialStack~\cite{spatialstack}, MiMo-Embodied~\cite{mimoembodied}, Embodied-R1.5~\cite{embodiedr15}, RoboBrain~2.5-4B~\cite{robobrain25}, and HY-Embodied-0.5~\cite{hyembodied05}. GeoSR3D~\cite{geosr}, GeoVR-4B~\cite{geovr}, and Spatial-MLLM-1.1~\cite{spatialmllm} use the same visual inputs as GT-VLM.

\supppair{%
\subsection{Evaluation Inputs and Sample Counts}
Table~\ref{tab:evalcounts} lists the benchmark releases used for evaluation. Following the \href{https://openaccess.thecvf.com/content/CVPR2025/supplemental/Yang_Thinking_in_Space_CVPR_2025_supplemental.pdf}{VSI-Bench sampling protocol} of Yang et al.~\cite{vsi}, we uniformly sample 32 frames per video and present them as an ordered sequence of images. CV-Bench, EmbSpatial, and MMBench-Dev use one image per item. The number of images per question ranges from 1 to 32 on ViewSpatial, 1 to 4 on BLINK, and 2 to 5 on All-Angles. When a question includes multiple images, we present them in the order provided by the benchmark.

VSI-Bench averages eight spatial task scores, using accuracy for multiple-choice questions and mean relative accuracy for numerical answers~\cite{vsi}. On the separately released \href{https://huggingface.co/datasets/jasonzhango/SPAR-Bench}{SPAR-Bench test set} (7,211 items), we report the mean of the 20 task scores~\cite{spar}. We extract the final answer and assign zero credit to empty selections. On MMBench-Dev, we report per-item accuracy with each item presented once in its supplied option order.

We use greedy decoding with thinking disabled for all variants in Table~\ref{tab:fullablation}. The maximum answer length is 1,000 tokens, or 1,024 for SITE; decoding stops at the end-of-sequence token. Failed generations count as incorrect.

\begingroup\captionsetup{type=table}
\centering\small
\begin{tabular}{lrrr}
\toprule
Benchmark & Items & Gen. errors & Parse failures \\
\midrule
CV-Bench~\cite{cvbench} & 2,638 & 0 & 0 \\
SPAR-Bench~\cite{spar} & 7,211 & 0 & 0 \\
MMSI-Bench~\cite{mmsi} & 1,000 & 0 & 1 \\
ViewSpatial-Bench~\cite{viewspatial} & 5,712 & 0 & 0 \\
All-Angles-Bench~\cite{allangles} & 2,132 & 0 & 0 \\
VSI-Bench~\cite{vsi} & 5,130 & 0 & 0 \\
EmbSpatial-Bench~\cite{embspatial} & 3,640 & 0 & 0 \\
SITE~\cite{site} & 8,068 & 0 & 0 \\
MMBench-Dev~\cite{mmbench} & 4,377 & 4 & 4 \\
BLINK Val All~\cite{blink} & 1,901 & 0 & 0 \\
\bottomrule
\end{tabular}
\caption{\textbf{Evaluation sample counts and generation outcomes.} The evaluations span single-image, multi-view, and video inputs. Counts and failures refer to GT-VLM c2; generation and parsing failures may overlap.}
\label{tab:evalcounts}
\par\endgroup
}{%
\subsection{CV-Bench and BLINK}
We use CV-Bench's official balanced accuracy, as defined by Tong et al.~\cite{cvbench} in the \href{https://huggingface.co/datasets/nyu-visionx/CV-Bench\#accuracy}{benchmark release}. Let $A_{\mathrm{ADE}}$, $A_{\mathrm{COCO}}$, and $A_{\mathrm{Omni}}$ denote source accuracies. CV-2D is $(A_{\mathrm{ADE}}+A_{\mathrm{COCO}})/2$, CV-3D is $A_{\mathrm{Omni}}$, and the reported CV overall is their equal-weight mean. For c2, CV-2D is 82.64 and CV-3D is 92.17, giving 87.40 overall.

BLINK Val All covers the 1,901 questions in the released validation set~\cite{blink}. GT-VLM scores 85.32 on the separate spatial diagnostic, which averages Multi-view Reasoning, Relative Depth, and Spatial Relation.
\subsection{SITE Chance Adjustment}
SITE~\cite{site} adjusts for the probability of guessing correctly, which varies with the number of options. The authors' \href{https://github.com/wenqi-wang20/SITE-Bench/blob/e1b85b38eb5e31b2023cba91f809e0e7b3fb1e8b/eval_scripts/aggregate.py}{evaluation script} pools the image and video partitions. With correctness $z_i\in\{0,1\}$ and $K_i$ choices for question $i$, the full-set score is
\begin{equation}
 \mathrm{CAA}=100\frac{\sum_i(z_i-1/K_i)}{\sum_i(1-1/K_i)}.
\end{equation}
GT-VLM answers 2,677 of 4,449 image items and 2,458 of 3,619 video items correctly, for a pooled CAA of 48.23\%. Answer extraction is case-sensitive and uses a seeded fallback for unparsed responses.

\subsection{Spatial and Overall Averages}
Spatial Avg. assigns weight $1/8$ to CV-Bench overall, SPAR-Bench, MMSI-Bench, ViewSpatial-Bench, All-Angles-Bench, VSI-Bench, EmbSpatial-Bench, and SITE CAA. This group covers 2D and 3D reasoning and integrated spatial tasks. Means are computed before display rounding.

Overall Avg. is $\bar{s}=\tfrac{1}{10}\sum_b s_b$ over the ten benchmark aggregates. The ten terms are the eight spatial-group benchmarks plus MMBench and BLINK All, each with equal weight.
}

\clearpage
\section{Geometry Interfaces and Local Compression}
\begin{table}[H]
\centering\small
\renewcommand{\arraystretch}{1.16}\begin{tabular*}{\linewidth}{@{\extracolsep{\fill}}lrrrrrrr@{}}
\toprule
 & \multicolumn{1}{c}{\textbf{GT-VLM}} & \multicolumn{2}{c}{Compression} & \multicolumn{4}{c}{Geometry interface} \\
\cmidrule(lr){2-2}\cmidrule(lr){3-4}\cmidrule(l){5-8}
Benchmark & c2 & c4 & c8 & Direct Fusion & Token Merge & PatchOnly & SpatialStack-style \\
\midrule
CV-Bench & 87.40 & 88.28 & 87.64 & 88.29 & 87.72 & 87.62 & 88.36 \\
SPAR-Bench & 69.51 & 69.21 & 68.63 & 68.72 & 68.68 & 69.03 & 69.07 \\
MMSI-Bench & 39.10 & 38.90 & 38.50 & 37.80 & 37.30 & 38.10 & 37.50 \\
ViewSpatial-Bench & 53.36 & 53.80 & 53.03 & 53.31 & 53.57 & 53.24 & 53.64 \\
All-Angles-Bench & 57.41 & 54.13 & 52.49 & 52.20 & 51.69 & 53.19 & 53.28 \\
VSI-Bench & 67.42 & 67.32 & 66.42 & 66.53 & 66.61 & 66.90 & 66.34 \\
EmbSpatial-Bench & 80.52 & 80.80 & 80.00 & 80.03 & 79.09 & 79.53 & 78.76 \\
SITE (CAA) & 48.23 & 47.93 & 47.25 & 47.86 & 48.22 & 48.16 & 47.74 \\
\addlinespace[3pt]
MMBench-Dev & 88.51 & 88.03 & 88.17 & 88.12 & 88.05 & 88.35 & 88.07 \\
BLINK Val All & 68.54 & 68.44 & 68.65 & 66.96 & 64.39 & 66.49 & 64.23 \\
\midrule
Spatial Avg. & \textbf{62.87} & 62.55 & 61.74 & 61.84 & 61.61 & 61.97 & 61.84 \\
Overall Avg. & \textbf{66.00} & 65.68 & 65.08 & 64.98 & 64.53 & 65.06 & 64.70 \\
\bottomrule
\end{tabular*}

\caption{\textbf{Complete interface and compression results (\%).} c2 has the highest ten-benchmark mean. Stronger compression preserves much of the aggregate performance but reduces All-Angles accuracy. Spatial Avg. covers the first eight benchmarks; Overall Avg. includes MMBench-Dev and BLINK Val All.}
\label{tab:fullablation}
\end{table}
\supppair{%
\paragraph{Local patch inputs.}
GT-VLM and Token Merge both use DPT and the native-feature bypass. GT-VLM gives each local geometry patch its own input position; Token Merge adds it to the corresponding RGB token. In both models, each frame's camera token and sixteen register tokens occupy seventeen separate positions in the language-model input. With dedicated local patch positions, GT-VLM scores 62.87 on Spatial Avg., compared with 61.61 for Token Merge.

For $F$ geometry frames, $V$ RGB tokens, and $T$ text and template tokens, Token Merge uses $T+V+17F$ input positions. GT-VLM uses one more position per compressed local patch. For example, a two-view All-Angles prompt has 135 RGB tokens and 135 local patches per view, plus 107 text and template tokens. These counts give 681 positions for GT-VLM and 411 for Token Merge.

\paragraph{Local and global geometry.}
PatchOnly keeps the local patch positions but omits the camera and register tokens. Restoring these tokens raises the spatial average from 61.97 to 62.87 (Table~\ref{tab:fullablation}).

}{%
\paragraph{Other interfaces.}
Direct Fusion projects and concatenates four geometry feature levels; the SpatialStack-style variant injects layers 11, 17, and 23 at decoder layers 0, 1, and 2. Both score 61.84. GT-VLM gains the most over all four interfaces on All-Angles, scoring 5.72 points above Token Merge and 4.13 above the SpatialStack-style variant.

\paragraph{Local compression.}
Moving from c2 to c4 raises CV-Bench accuracy from 87.40 to 88.28 and ViewSpatial from 53.36 to 53.80. All-Angles shows the opposite pattern: accuracy falls from 57.41 at c2 to 54.13 at c4 and 52.49 at c8. We keep c2 to preserve performance on these cross-view questions.
}

\clearpage
\section{Geometry Intervention Protocol}
After assembling the input sequence, we replace geometry embeddings with those from another scene or set them to zero. We apply each intervention to both branches together or to just the local patches or the global camera/register tokens. Only the selected embeddings change: RGB and text, token IDs and positions, modality tags, attention masks, and sequence length stay fixed.

\supppair{%
We assign a donor from a different scene to each recipient using the released video identity, with recognized same-scene capture suffixes grouped together. Donor inputs have distinct decoded content and match the recipient in frame count, grid dimensions, padding, feature shape, and dtype. We fix the cyclic donor assignment before generation using a seeded ordering. All conditions use the same 5,130 questions and 288 recipient scenes.

Replacing both branches with geometry from another scene lowers relative-direction accuracy from 79.99 to 70.55, while the object-size score is nearly unchanged (72.63 vs.\ 72.75; Table~\ref{tab:interventiontasks}).
}{%
\begingroup\captionsetup{type=table}
\centering\small
\begin{tabular}{lrrr}
\toprule
Task & Matched & Wrong scene & Zero \\
\midrule
Appearance order & 77.83 & 75.40 & 67.31 \\
Absolute distance & 49.39 & 44.41 & 43.12 \\
Object counting & 73.19 & 71.19 & 58.02 \\
Relative direction & 79.99 & 70.55 & 64.20 \\
Relative distance & 66.62 & 65.35 & 60.85 \\
Object size & 72.63 & 72.75 & 70.93 \\
Room size & 69.72 & 63.65 & 63.72 \\
Route planning & 50.00 & 44.85 & 40.21 \\
\bottomrule
\end{tabular}

\caption{\textbf{VSI task scores for the whole-geometry conditions in Table~\ref{tab:intervention}.} Relative direction and object counting show the largest drops when both local and global geometry embeddings are zeroed.}
\label{tab:interventiontasks}
\par\endgroup
}

Paired uncertainty uses 2,000 recipient-scene bootstrap replicates with seed 3047. Each bootstrap sample contains all questions and paired conditions from its selected scenes, and the VSI task/direction aggregate is recomputed. The checkpoint and donor map are held fixed, so these intervals describe test-scene variation rather than variation across training runs or donor assignments. Task-specific intervals are unadjusted for multiple comparisons.

Intervention differences are measured against the scene-matched control, which scores 67.42.

\paragraph{Branch-specific results.}
With the same protocol, wrong-scene patches score 66.60 with a paired 95\% difference interval of $[-1.33,-0.23]$, while zero patches score 61.11 with interval $[-7.54,-4.89]$. Camera/register wrong-scene and zero conditions score 63.28 and 63.29, with intervals $[-5.21,-2.88]$ and $[-5.33,-2.85]$, respectively.

\section{Manipulation Setup and Perturbation Results}
Original Qwen3.5, instruction-tuned Qwen3.5, and GT-VLM c2 use the same GR00T~\cite{groot} action head architecture and initialization procedure. We train the head and language decoder with the same recipe within each benchmark. In GT-VLM, we also train the geometry tokenizer but keep the VGGT-Omega geometry encoder frozen. At rollout, GT-VLM predicts actions using the current observation's local patch tokens and global camera and register tokens.

For each model, we train a single policy for seven epochs on all 2,000 original demonstrations from the four LIBERO suites~\cite{libero}: Spatial, Object, Goal, and LIBERO-10. We report mean success across the four suites (ten tasks per suite, fifty trials per task). On RoboTwin clean-to-clean~\cite{robotwin,robotwin2}, we train on 2,500 clean demonstrations for 30K steps and evaluate on fifty tasks with one hundred clean trials each.

\supppair{%
On LIBERO-Plus~\cite{liberoplus}, we evaluate each model's LIBERO checkpoint without further training. The \href{https://github.com/sylvestf/LIBERO-plus}{official protocol} uses one trial per perturbed task variant, for 10,030 trials in total. Overall success is the fraction of successful trials across all seven categories (Table~\ref{tab:perturbationfull}).

\paragraph{Public manipulation references.}

The OpenVLA~\cite{openvla} LIBERO score averages three seeds, and OpenVLA-OFT~\cite{openvlaoft} uses independently trained policies for the four suites. The $\pi_0$ and $\pi_{0.5}$ LIBERO scores come from the released 30K-step OpenPI policies~\cite{openpiref,openpi05ref}. X-VLA~\cite{xvla} and Xiaomi Robotics-0~\cite{xiaomirobotics0} report four-suite LIBERO results and jointly trained RoboTwin policies. For RoboTwin c2c~\cite{robotwinboard}, $\pi_0$ is fine-tuned per task and $\pi_{0.5}$ is trained jointly across fifty tasks.
}{%
\begingroup\captionsetup{type=table}
\centering\small
\begin{tabular*}{\linewidth}{@{\extracolsep{\fill}}lrrrrr@{}}
\toprule
Perturbation & Episodes & Base & FT & GT-VLM & $\Delta$ \\
\midrule
Camera view & 1599 & 41.65 & 56.97 & 63.91 & 6.94 \\
Sensor noise & 1601 & 78.33 & 86.38 & 84.63 & -1.75 \\
Object layout & 1525 & 80.98 & 81.44 & 82.95 & 1.51 \\
Robot initial state & 1550 & 49.16 & 52.77 & 51.87 & -0.90 \\
Lighting & 1142 & 95.36 & 94.75 & 95.88 & 1.14 \\
Background texture & 1076 & 95.17 & 93.12 & 94.24 & 1.12 \\
Language instruction & 1537 & 84.58 & 83.02 & 87.25 & 4.23 \\
\midrule
Weighted total & 10,030 & 73.08 & 76.91 & 78.72 & 1.81 \\
\bottomrule
\end{tabular*}

\caption{\textbf{LIBERO-Plus category success rates (\%).} Base starts from the original Qwen3.5-4B; FT starts from Qwen3.5-4B finetuned on our instruction data. $\Delta$ is GT-VLM minus FT in percentage points; the largest gain is under camera-view changes.}
\label{tab:perturbationfull}
\par\endgroup
}

\clearpage
\section{RTX 5090 Inference Profiling}
Each model makes one inference pass over all 2,638 CV, 2,132 All-Angles, and 5,130 VSI items on one RTX 5090 at batch size one. Timing statistics exclude the first two warmup requests per job and decoder-cold records, giving 2,636, 2,130, and 5,128 observations respectively. The profile uses merged LoRA, BF16, greedy non-thinking decoding, the native end-of-sequence behavior, a 1,000-token cap, and KV caching. Native CUDA graphs are enabled, and each request computes its own image features and KV cache.

\begin{table}[H]
\centering\small
\setlength{\tabcolsep}{3pt}
\renewcommand{\arraystretch}{1.12}\begin{tabular*}{\linewidth}{@{\extracolsep{\fill}}llrrrrrrrr@{}}
\toprule
 & & & \multicolumn{4}{c}{Prefill components} & \multicolumn{3}{c}{Measured intervals} \\
\cmidrule(lr){4-7}\cmidrule(l){8-10}
Set & Model & Timed $n$ & 2D & 3D core & LLM & Other & Prefill & Post-first & Total \\
\midrule
CV & Base & 2636 & 8.00 & 0.00 & 28.10 & 4.96 & 42.86 & 52.90 & 107.48 \\
CV & c2 & 2636 & 7.41 & 36.44 & 42.54 & 2.21 & 95.24 & 25.95 & 121.16 \\
CV & c4 & 2636 & 7.36 & 36.40 & 26.77 & 2.20 & 86.71 & 25.83 & 112.55 \\
\midrule
Angles & Base & 2130 & 7.36 & 0.00 & 26.33 & 4.91 & 38.63 & 2276.18 & 2320.41 \\
Angles & c2 & 2130 & 7.37 & 23.72 & 42.52 & 2.53 & 76.10 & 8.82 & 85.12 \\
Angles & c4 & 2130 & 7.37 & 23.72 & 26.92 & 2.48 & 60.48 & 42.79 & 120.46 \\
\midrule
VSI & Base & 5128 & 45.39 & 0.00 & 91.88 & 8.52 & 146.84 & 998.45 & 1134.17 \\
VSI & c2 & 5128 & 44.65 & 152.23 & 191.36 & 16.62 & 405.69 & 28.10 & 435.34 \\
VSI & c4 & 5128 & 43.36 & 146.26 & 140.16 & 11.40 & 341.36 & 27.47 & 376.10 \\
\bottomrule
\end{tabular*}

\caption{\textbf{Median inference times on RTX 5090 (ms).} Total is complete-answer wall-clock time; component medians are not additive.}
\label{tab:costfull}
\end{table}

Total prefill is the CUDA interval from model start through first logits. It includes the Qwen visual encoder (2D), the per-request sum of VGGT, DPT, and camera/register adaptation (3D core), and LLM prefill. Other prefill is the per-request residual, covering packing, padding, scatter, positions, the first output head, and launch gaps. We sum the three core intervals for each request before taking their median. Table~\ref{tab:costfull} gives the component medians alongside measured prefill and complete-answer times.

Post-first is the CUDA interval from first logits to natural generation end, including sampling, subsequent autoregressive decoding, and launch gaps. Total latency is independently measured wall-clock time from generation start to natural completion. CPU media reading, video decoding, preprocessing, H2D, and explicit capture/reference/warmup are excluded.

\begin{table}[H]
\centering\small
\begin{tabular*}{\linewidth}{@{\extracolsep{\fill}}lrrrrrr@{}}
\toprule
& \multicolumn{3}{c}{Median output length (tokens)} & \multicolumn{3}{c}{Peak allocated memory (GiB)} \\
\cmidrule(lr){2-4}\cmidrule(l){5-7}
Benchmark & Base & c2 & c4 & Base & c2 & c4 \\
\midrule
CV & 7.00 & 4.00 & 4.00 & 9.168 & 14.089 & 14.089 \\
All-Angles & 267.00 & 2.00 & 6.00 & 9.173 & 11.839 & 11.842 \\
VSI & 112.50 & 4.00 & 4.00 & 9.380 & 12.455 & 12.462 \\
\bottomrule
\end{tabular*}

\caption{\textbf{Output length and peak memory on RTX 5090.} On All-Angles, c2 generates shorter answers than c4 despite similar peak memory use.}
\label{tab:costresources}
\end{table}

On All-Angles, c2 answers sooner than c4 despite a slower prefill. Its median answer is two tokens rather than six, and it spends 8.82\,ms after the first logits compared with 42.79\,ms for c4. On VSI, both models generate a median of four tokens, and their post-first times are similar: 28.10\,ms for c2 and 27.47\,ms for c4. Base generates longer answers on both benchmarks.

\end{document}